\documentclass{article}
\usepackage{ijcai26}

\usepackage{times}
\usepackage{soul}
\usepackage{url}
\usepackage[hidelinks]{hyperref}
\usepackage[utf8]{inputenc}
\usepackage[small]{caption}
\usepackage{graphicx}
\usepackage{amsmath}
\usepackage{amsthm}
\usepackage{booktabs}
\usepackage{algorithm}
\usepackage{algorithmic}
\usepackage[switch]{lineno}
\usepackage{natbib}
\usepackage{multirow}
\usepackage{tabularx}
\usepackage{array}

\title{Component-Level Lesioning of Language Models Reveals Clinically Aligned Aphasia Phenotypes}

\author{
Yifan Wang$^{1*}$
\and
Jichen Zheng$^{2,3*}$\and
Jingyuan Sun$^{1*}$\and
Yunhao Zhang$^{2,3}$\and
Chunyu Ye$^{2,3}$\and
Jixing Li$^{4}$\and
Chengqing Zong$^{2,3}$\and
Shaonan Wang$^5$\\
\affiliations
$^1$Department of Computer Science, The University of Manchester, UK\\
$^2$State Key Laboratory of Multimodal Artificial Intelligence Systems, Institute of Automation, CAS, Beijing, China\\
$^3$School of Artificial Intelligence, University of Chinese Academy of Sciences, Beijing, China\\
$^4$Department of Linguistics and Translation, City University of Hong Kong, Hong Kong\\
$^5$Department of Language Science and Technology, Hong Kong Polytechnic University, Hong Kong\\
\emails
yifan.wang-38@postgrad.manchester.ac.uk,
zhengjichen2023@ia.ac.cn,
jingyuan.sun@manchester.ac.uk,
\{zhangyunhao2021, yechunyu2023\}@ia.ac.cn,
jixingli@cityu.edu.hk,
cqzong@nlpr.ia.ac.cn,
shaonan.wang@polyu.edu.hk
}

\begin{document}

\maketitle

\begin{abstract}
Large language models (LLMs) increasingly exhibit human-like linguistic behaviors and internal representations that they could serve as computational simulators of language cognition. We ask whether LLMs can be systematically manipulated to reproduce language-production impairments characteristic of aphasia following focal brain lesions. Such models could provide scalable proxies for testing rehabilitation hypotheses, and offer a controlled framework for probing the functional organization of language.
We introduce a clinically grounded, component-level framework that simulates aphasia by selectively perturbing functional components in LLMs, and apply it to both modular Mixture-of-Experts models and dense Transformers using a unified intervention interface. Our pipeline (i) identifies subtype-linked components for Broca’s and Wernicke’s aphasia, (ii) interprets these components with linguistic probing tasks, and (iii) induces graded impairments by progressively perturbing the top-k subtype-linked components, evaluating outcomes with Western Aphasia Battery (WAB) subtests summarized by Aphasia Quotient (AQ). Across architectures and lesioning strategies, subtype-targeted perturbations yield more systematic, aphasia-like regressions than size-matched random perturbations, and MoE modularity supports more localized and interpretable phenotype-to-component mappings. These findings suggest that modular LLMs, combined with clinically informed component perturbations, provide a promising platform for simulating aphasic language production and studying how distinct language functions degrade under targeted disruptions.

\end{abstract}

\section{Introduction}
Large language models (LLMs) have advanced rapidly, achieving strong performance across a wide range of language tasks. Beyond task accuracy, they increasingly reproduce human-like linguistic behaviors \citep{park2023generative, mou2024individual} and exhibit internal representations that align with human neural responses during language processing \citep{schrimpf2021neural, caucheteux2022brains,toneva2019interpreting, goldstein2024alignment}. These developments position LLMs not only as engineering artifacts, but also as candidate computational models for investigating the mechanisms of human language—and, crucially, for simulating disordered language in a controlled and scalable way.

Aphasia is an especially compelling testbed for this goal. As a heterogeneous language disorder caused by focal neural injury, aphasia presents with dissociable impairments in comprehension and/or production that vary systematically across subtypes. If we can manipulate an LLM’s internal components to reproduce the linguistic profiles observed in patients—capturing both the type and severity of deficits—then LLMs could serve as computational proxies for individuals with aphasia. Such proxies would enable efficient, reproducible testing of therapeutic hypotheses and could support the development of personalized rehabilitation strategies. More broadly, they would offer a new paradigm for probing functional organization in both biological and artificial language systems.

The scientific motivation for this approach draws on the long-standing view that language is supported by a modular or partially specialized architecture in the brain \citep{dronkers2023neuroscience, meunier2010modular, bertolero2015modular}. Under this framework, different neural substrates contribute different linguistic functions, and damage to particular regions yields relatively predictable patterns of impairment. Classic descriptions of Broca’s and Wernicke’s aphasias, for example, link deficits in speech production and comprehension to lesions in distinct cortical territories. Early computational accounts pursued this logic using small-scale connectionist models, showing that “lesioning” specific parts of a network could reproduce select patient-like error patterns \citep{dell1997lexical, farah1991cognitive, hinton1991lesioning}. While foundational, these models lacked the scale and linguistic expressivity needed to approximate the breadth and complexity of human language, limiting their value as realistic proxies for aphasic syndromes.

In this paper, we leverage modular LLMs based on a Mixture-of-Experts (MoE) architecture, where different experts often capture different types of information—suggesting a form of functional specialization analogous to that observed in the human brain \citep{chen2022task, zhang2023emergent}. We also study dense Transformer models as baselines. To enable a fair comparison, we treat all intervention targets as functional components: in MoE models, a component is an expert, whereas in dense models, a component corresponds to a fine-grained neuron group or feature channel within a layer. We first identify components most associated with two clinically salient aphasia subtypes, Broca’s and Wernicke’s. We then interpret these components using linguistic probing tasks. Finally, we induce graded impairments by perturbing the top-2\% subtype-linked components and evaluate the resulting deficits using Western Aphasia Battery (WAB) subtests summarized by the Aphasia Quotient (AQ), a standard clinical measure. Across experiments, progressive impairment yields graded, clinically measurable declines on WAB subtests and AQ. Together, these results suggest that LLMs—particularly modular architectures—can serve as scalable experimental platforms for modeling aphasia, enabling controlled studies of how distinct language functions degrade under targeted perturbations.

\section{Related Work}

Our research integrates insights from three distinct but converging domains: the neuroscience of aphasia, the history of computational disorder simulation, and the architecture of modern language models. A central gap is that most recent “LLM damage” studies emphasize general robustness, but do not align to standardized clinical scoring or separate clinically meaningful aphasia subtypes.

\subsection{Aphasia and the Neural Basis of Language}
Aphasia offers a clinically grounded lens on the functional organization of language. Broca’s aphasia is typically associated with effortful, agrammatical production with relatively preserved comprehension, whereas Wernicke’s aphasia is characterized by fluent but semantically disrupted speech \citep{lichtheim1885aphasia, damasio1992aphasia}. Lesion and neuroimaging studies support partially dissociable substrates: the left inferior frontal gyrus (IFG) is consistently linked to syntactic processing \citep{friederici2011brain}, while temporal regions contribute to semantic integration \citep{dronkers2004lesion, friederici2011brain}. These findings motivate a computational analogue where localized perturbations produce interpretable linguistic deficits and subtype-specific profiles.

\subsection{Simulating Language Disorders with Computational Models} 
Lesion-based simulation has a long tradition in cognitive modeling. Early connectionist models reproduced naming errors and grammatical impairments by selectively damaging model components \citep{joanisse1999impairments, thomas2002modelling}. More recently, LLMs have been used to elicit aphasia-like outputs via controlled prompting \citep{manir2024llm} or by perturbing high-level structures such as layers \citep{wang2025emergent}. While promising, these approaches are often coarse-grained: they typically assess degradation with broad Natural Language Processing (NLP) benchmarks, making it difficult to connect a perturbed component to a specific linguistic function, and rarely provide standardized clinical evaluation or clear subtype separation. Our work builds on this tradition but seeks a model with a more direct and biologically plausible analogy to the brain's modularity.

\subsection{Mixture-of-Experts Models and Component-Level Lesioning}
Mixture-of-Experts models activate only a subset of parameters (experts) per token \citep{shazeer2017outrageously,fedus2022switch,jiang2024mixtral}. Beyond efficiency, experts can develop emergent specialization for tasks or linguistic phenomena \citep{chen2022task,zhang2023emergent}, which makes MoE architectures well-suited for component-level lesion studies. However, the broader LLM impairment literature has mainly focused on general capability loss, safety, or robustness \citep{sharma2023truth, wang2023robustness, gong2025knowledge}, and often lacks clinical alignment and subtype-sensitive analysis. It is also unclear whether the apparent locality of MoE lesions reflects functional modularity or is partly shaped by MoE routing.

To address these issues, we study both MoE and dense Transformer baselines under a unified notion of functional components (experts in MoE; neuron groups/feature channels in dense models). This allows architecture-independent perturbation and direct comparison, while clinical evaluation provides a standardized target for assessing subtype-specific impairments.

\begin{figure*}[t]
    \centering
    \includegraphics[width=\textwidth]{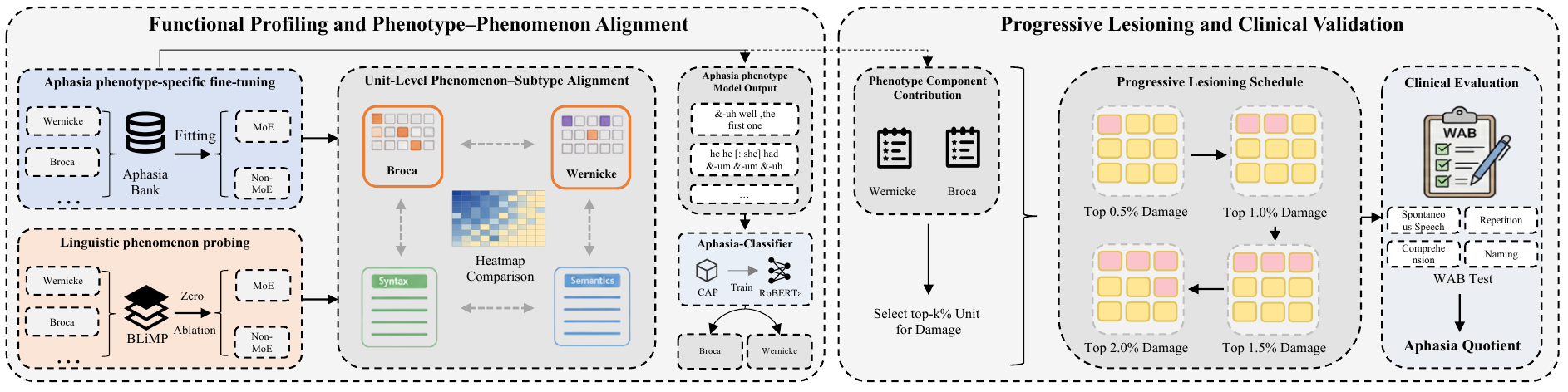}
    \caption{\textbf{Details of the analysis pipeline}. \textbf{Functional Profiling and Phenotype–Phenomenon Alignment:} BLiMP probing and AphasiaBank fine-tuning yield unit-importance rankings; a heatmap links linguistic phenomena to Broca/Wernicke units, with CAP as an external phenotype check. \textbf{Progressive Lesioning and Clinical Validation:} Top-k\% units are progressively lesioned and evaluated on WAB/AQ, enabling a matched comparison between MoE and dense models.}
    \label{fig:flowchart}
\end{figure*}

\section{Method}
We apply the framework to two transformer LMs: a dense baseline (OLMo) and its Mixture-of-Experts variant (OLMoE). To make the two models comparable, all interventions operate on a shared notion of functional components (units)—a layer–expert pair in OLMoE, and a hidden dimensions in the Feedforward Neural Network (FFN) for OLMo. The methodology follows a single pipeline for both models: (1) attribute fine-grained linguistic phenomena to units using BLiMP, (2) identify phenotype-linked units from AphasiaBank fine-tuning signals and validate subtype separability with an external CAP classifier, (3) select a stable top-$p\%$ threshold via p-sweep, (4) align phenotype-linked units with linguistic phenomena through a rank-percentile heatmap, and (5) induce graded impairments by progressively lesioning the top-ranked units and quantify clinical degradation with Western Aphasia Battery (WAB), summarized as the Aphasia Quotient (AQ) (Figure \ref{fig:flowchart}). 

\subsection{Model Architecture}
We use OLMoE \citep{muennighoff2024olmoe}, specifically OLMoE-1B-7B-0924-Instruct. It keeps a standard Transformer backbone but replaces the FFN with MoE layers, where a lightweight router selects 8 of 64 experts per token. The model has 16 layers, activates about 1B parameters per forward pass, and is fine-tuned on OLMo-Instruct.

As a non-MoE control, we use the matched dense OLMo-\textsc{Instruct} checkpoint \citep{groeneveld2024olmo}. OLMo has the same tokenizer, depth, and training corpus as OLMoE, but uses a conventional dense FFN in each layer with no routing or sparse expert activation, allowing us to isolate the effect of MoE-style modularity. These two models therefore support a controlled architectural comparison under the same unit-level probing and lesion interface introduced next.

\subsection{Unit Attribution via BLiMP Zero-Ablation}
\label{sub:3.2}
This section produces a task-wise unit attribution map that links internal units to specific linguistic phenomena. We use BLiMP’s minimal-pair evaluation to estimate the causal importance of model units for specific linguistic phenomena. For each BLiMP subtask, we ablate one unit at a time by zeroing its output and record the resulting accuracy. In OLMoE, the ablated unit is a single expert, with the router and all other experts left unchanged. In the dense OLMo baseline, we treat FFN hidden dimensions within each layer as parallel neuron groups and apply the same output-zeroing intervention, enabling a comparable unit-level analysis across architectures.

We define the importance of an expert's specific task as the change in accuracy:
\begin{equation}
\label{equ:1}
\Delta_t(u) = Acc_t(u) - Acc_t
\end{equation}
Negative $\Delta_t(u)$ indicates that ablation harms performance, suggesting that model expert or neurons $(u)$ contributes positively to phenomenon $t$. Repeating this procedure yields a task-specific attribution map $\Delta_t \in \mathbf{R}^{u}$ for each BLiMP sub-task. These BLiMP-derived rankings are later used to compute rank-percentiles for the alignment heatmaps and to construct task profiles for threshold robustness analysis.

\subsection{Aphasia Phenotype-Specific Unit Contribution}
\label{sub:3.3}
While Section \ref{sub:3.2} attributes phenomena to units, this section attributes clinical phenotypes (Broca vs. Wernicke) to units using AphasiaBank supervision. The BLiMP analysis links units to specific syntactic/semantic phenomena. To identify which units are most involved in fitting aphasia subtypes, we fine-tune the same base model (OLMoE or OLMo) on the Broca and Wernicke subsets of AphasiaBank, producing two phenotype-specific models. We then rank units by their training-time contribution using gradient-based statistics aggregated at the unit level (experts in OLMoE; neuron groups in OLMo), following \citet{zhang2024unveiling}.

For phenotype $c \in {Broca, Wernicke}$, we define the cumulative gradient–weight importance for a parameter tensor $\Theta$ as:
\begin{equation}
\label{equ:2}
I_c(\theta) = \sum_{s=1}^{S} \left|\, g_{\theta}^{(s)} \odot \theta^{(s)} \,\right|
\end{equation}
where $\odot$ is element-wise multiplication. $\theta^{(s)}$ is the weight at step $s$, and $g_{\theta}^{(s)}$ is the gradient of this weight at step $s$. We then aggregate this quantity over all parameters belonging to a component $u$ to obtain a component-level contribution score:
\begin{equation}
\label{equ:3}
\mathrm{Score}_c(u) = \sum_{\theta \in \Theta(u)} \sum I_c(\theta)
\end{equation}
Ranking experts by $\mathrm{Score}_c(u)$ yields two phenotype-specific component contribution tables (Broca and Wernicke), which we use for downstream alignment with BLiMP and progressive lesion experiments.

To check that the fine-tuned models produce subtype-consistent language rather than simply memorizing aphasia-related content, we add an external style validator. A lightweight RoBERTa-base \citep{liu2019roberta} classifier is trained on TalkBank’s Comparative Aphasia Project English data (CAP) \citep{cap_talkbank, bates1989comparative} to predict whether a sentence is more Broca-like or Wernicke-like, returning subtype probabilities from text alone. We run this classifier on model generations and report style consistency as the fraction of outputs assigned to the intended subtype, providing an independent corpus-based verification that the two fine-tuned models are separable in phenotype. Together, the phenotype-linked unit ranking and the CAP-based separability check provide the lesion candidates and validation signals for the subsequent threshold selection and alignment analyses.

\subsection{Threshold Selection and Robustness}
Because both visualization (alignment heatmaps) and progressive lesioning require selecting the top-$p\%$ units, a robustness protocol is needed to avoid arbitrary threshold choices. To keep the analysis comparable across MoE and dense models, all ablatable components are treated as \emph{units}. For each BLiMP subtask, units are converted to rank-percentiles based on the attribution scores from Section \ref{sub:3.2}. For each subtype (Broca/Wernicke), units are ranked by the gradient-based contribution scores (Section \ref{sub:3.3}) and the top-$p\%$ are selected. Since $p$ acts as a visualization threshold, we run a p-sweep over $p \in {0.5, 1, 2, 3, 5, 10}\%$: for each $p$, we compute a BLiMP task-profile vector by averaging the selected units’ rank-percentiles within each subtask, and then measure stability across thresholds using Spearman correlation:
\begin{equation}
\label{equ:4}
\rho(p_1, p_2) = Spearman(m_{p1}, m_{p2})
\end{equation}
If $\rho(p_1, p_2)$ is consistently high within the range of $p$ mentioned above, it indicates that the observed cross-task hotspot pattern is not sensitive to the threshold selection. For readability, we will choose an intermediate threshold for visualization. In the following analyses, an intermediate $p=2\%$ is adopted for visualization and lesion budgets, supported by the observed stability under p-sweep.

\subsection{Alignment of Language Phenomena with Clinical Phenotypes}
With the BLiMP unit rankings (Section \ref{sub:3.2}) and the subtype-linked contribution tables from AphasiaBank fine-tuning (Section \ref{sub:3.3}), the alignment is made explicit using a rank-percentile heatmap. For each subtype $c \in {\text{Broca}, \text{Wernicke}}$, the top-$p\%$ units are selected by their contribution scores. For each BLiMP subtask $t_j$, units are ordered by their BLiMP importance ranking, giving the rank $r_{t_j}(u_i)$ for each selected unit $u_i$. These ranks are then normalized by the number of units in that subtask to form a percentile matrix, where each entry indicates how strongly a subtype-linked unit is associated with a specific linguistic phenomenon.
\begin{equation}
\label{equ:5}
H_c(i,j) = \frac{r_{t_j}(u_i)}{N_{t_j}}
\end{equation}
$N_{t_j}$ represents the number of components in subtask $t_j$. A lower $H_c(i,j)$ indicates that the component is more important in the BLiMP subtask, and the closer it is to 1, the lower its importance. We visualized the matrix as a heatmap to compare the distribution of Broca syndrome-related experts and Wernicke syndrome-related experts on syntactic and semantic subtasks.

\subsection{Damage Model for Progressive Aphasia Lesions}
To mimic focal, lasting damage, we disable selected units while keeping the overall network structure intact. In OLMoE this targets experts; in dense OLMo it targets the corresponding FFN neuron groups. Two lesion schemes are compared: activation zeroing, which forces the unit’s output to zero, and Xavier re-initialization \citep{glorot2010understanding}, which overwrites the unit’s weights to erase its learned computation. In the Xavier-based lesion step, we traverse the selected units and replace their weight parameters with new values sampled from the Xavier uniform distribution, as formalized in Equation \ref{equ:6}:
\begin{equation}
\label{equ:6}
a = \sqrt{\frac{6}{n_{in} + n_{out}}}
\end{equation}
$n_{in}$ represents the number of input neurons, and $n_{out}$ represents the number of output neurons. We overwrite the original weights with a random value with a mean of 0 and a moderate variance, thereby completely destroying the original function and rendering it unable to perform its specific task.

Activation zeroing implements a strict silencing lesion by removing a unit’s output, whereas Xavier re-initialization creates a scrambling lesion by overwriting its weights, disrupting the learned computation while keeping the rest of the network intact. Together, they span two complementary impairment modes that better mirror clinical observations where damaged tissue may remain metabolically active yet functionally degraded \citep{kiran2019neuroplasticity, gleichgerrcht2015preservation, wilson2020neuroplasticity}. To model graded severity, units are ranked by the phenotype-linked contribution scores (Section \ref{sub:3.3}) and the top-$p\%$ are lesioned with increasing $p$; the resulting models are then evaluated with WAB in Section \ref{sub:3.7}.

\subsection{WAB-Based Clinical Evaluation and Aphasia Quotient}
\label{sub:3.7}
After constructing graded lesions on phenotype-linked units, Western Aphasia Battery (WAB) \citep{kertesz2007western} provides a standardized clinical endpoint to test whether targeted perturbations translate into aphasia-level functional loss. To assess whether our injury experiment would induce aphasia-like language changes, we evaluate models with four WAB subtests—\textbf{Spontaneous Speech (SS)}, \textbf{Comprehension (C)}, \textbf{Repetition (R)}, and \textbf{Naming (N)}—which probe fluency, understanding, repetition, and lexical retrieval. Each item is formatted as a standardized text prompt, and all models are tested with the same question set, prompting templates, and decoding settings; only minimal post-processing is applied.

Scores from the four subtests are combined into the \textbf{Aphasia Quotient (AQ)}. Since some WAB items require multimodal input or clinician-driven procedures that text-only models cannot perform, we keep only text-equivalent items (See the supplementary material wab-text.json file for details.). Each subtest is scored on this executable subset, normalized by the subset maximum, and then linearly rescaled to the original WAB subtest maximum to keep subtests comparable.
\begin{equation}
\label{equ:7}
AQ = (SS + \frac{C}{12} + \frac{R}{10} + \frac{N}{4}) \times 2
\end{equation}
We used the WAB standard weighting for the four subtests to derive an AQ score on a 0-100 scale. According to WAB documentation \citep{barfod2013western}, a score above 93.8 is considered normal, while a score below 93.8 is diagnosed as aphasia. We tested the AQ scores of a baseline model (no impairment) and a progressive impairment model, and observed how the AQ score changed with increasing impairment severity. We used AQ as our primary outcome measure for assessing the severity of aphasia.

\section{Experimental Setup}
\subsection{Dataset}
In this paper, we use four datasets:

\begin{itemize}
\item \textbf{Benchmark of Linguistic Minimal Pairs:} BLiMP \citep{warstadt2020blimp} evaluates grammatical knowledge using minimal sentence pairs. We use its syntax, semantics, and syntax--semantics subsets, and group examples into nine issue types using the \texttt{linguistics term} field.
\item \textbf{Comparative Aphasia Project English Data (CAP):} CAP is an English aphasia corpus released via TalkBank, with transcribed speech from 11 participants collected under a shared elicitation protocol for comparative analyses of aphasic language \citep{cap_talkbank, bates1989comparative}.
\item \textbf{Aphasia Bank:} AphasiaBank \citep{macwhinney2011aphasiabank} is a corpus of conversations, narratives, and Q\&A sessions from speakers with aphasia. To simulate Broca’s and Wernicke’s aphasia, 313 Broca samples and 63 Wernicke samples were selected.
\item \textbf{Western Aphasia Battery (WAB):} WAB is a widely used standardized clinical test for assessing language function in aphasia. It quantifies subtype and severity, and summarizes core oral subtests into a 0–100 Aphasia Quotient (AQ) \citep{kertesz2007western}.
\end{itemize}

\subsection{Evaluation Metrics}
In the evaluation of BLiMP language phenomena, we compute accuracy based on average log probability. Each sample contains a grammatically correct and incorrect sentence. If the model assigns higher average log probability to the correct sentence, it is counted as accurate.

In the style fidelity of the CAP classifier, we use a RoBERTa-based binary classifier. We report average classifier confidence (the predicted probability of the expected subtype). These metrics provide a corpus-based auxiliary validation demonstrating that phenotypic fine-tuning can produce discriminative subtype output patterns.

In the WAB clinical assessment, we consider four core subtests when evaluating the model—spontaneous speech, comprehension, repetition, and naming—and combine them into the 0-100 Aphasia Quotient (AQ) (see section \ref{sub:3.7} for details).

\subsection{Implementation Details}
For BLiMP zero-ablation, inference-only evaluation is conducted with activation patching using DeepSpeed on NVIDIA A100-SXM-64GB GPUs (global batch size 256; seed 42). The CAP-based subtype classifier is a RoBERTa-base sequence classifier fine-tuned with DeepSpeed and AdamW (lr 2e-5) for 10 epochs (batch size 16; seed 42). AphasiaBank model training and gradient-based attribution use DeepSpeed with AdamW (lr 5e-5) for 1 epoch (batch size 8; seed 1234), with ZeRO Stage-3 enabled for memory efficiency.

\section{Results and Discussion}
This section presents empirical evidence for the proposed component-level aphasia framework in both model architectures. It first tests whether subtype-linked components map onto fine-grained linguistic phenomena, providing an interpretable bridge between clinical phenotypes and behavioral signatures. It then motivates the choice of the top-p\% threshold with a p-sweep robustness analysis. Finally, it examines whether progressively lesioning these components yields graded declines on WAB/AQ, and illustrates the effects with qualitative examples.

\subsection{Style Consistency (CAP Classifier)}
To evaluate whether the model can retain the expected aphasia phenotype beyond task scores after fitting a corpus of real aphasia patients, we introduce an independent subtype consistency test based on a supervised classifier (RoBERTa base model) trained on the CAP dataset. 

As shown in Table \ref{tab:tb1}, the CAP classifier achieved an overall consistency of 98.25\% (OLMoE) and 92\% (OLMo) on the CAP validation set. The OLMoE outputs exhibited strong phenotypic consistency (85\% for the Broca subtype and 90\% for the Wernicke subtype). Notably, the non-MoE baseline model OLMo also achieved similarly high consistency (80\% for the Broca subtype and 95\% for the Wernicke subtype). This indicates that the observed aphasia subtype consistency is not an artifact caused by the MoE routing itself. In summary, these results establish that subtype-specific fine-tuning yields separable outputs, enabling downstream lesion selection and clinical evaluation.

\begin{center}
  \renewcommand{\arraystretch}{1.15}
  \setlength{\tabcolsep}{6pt}
  \begin{tabular}{lcc}
    \toprule
    & \textbf{OLMoE-output} & \textbf{OLMo-output} \\
    \midrule
    CAP Classifier Score & 98.25\% & 92\% \\
    Broca              & 85\%    & 80\% \\
    Wernicke           & 90\%    & 95\% \\
    \bottomrule
  \end{tabular}
  \captionof{table}{\textbf{CAP - classifier scores} on model outputs. \textbf{CAP Classifier Score} is evaluated on the held-out CAP validation split. \textbf{Broca/Wernicke} are subtype consistency of model generations that is the fraction of generations classified as the target subtype.}
  \label{tab:tb1}
\end{center}

\begin{figure}[!t]
  \centering
  \includegraphics[width=\columnwidth]{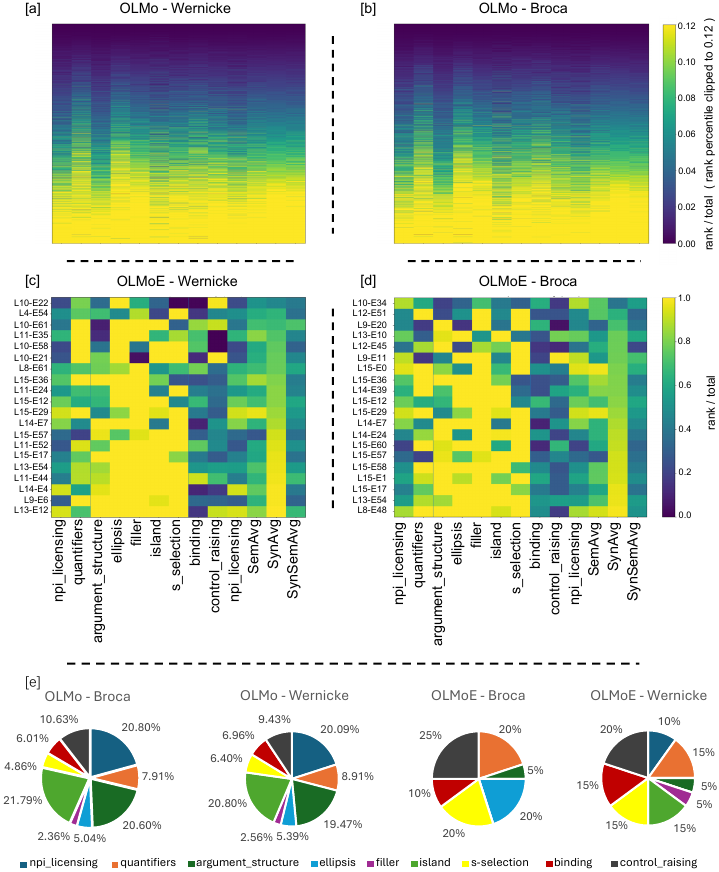}
  \caption{\textbf{Rank-percentile heatmaps and pie chart for top-2\% subtype-relevant units}. \textbf{[a–b] Dense (OLMo)} neurons (Wernicke, Broca). \textbf{[c–d] MoE (OLMoE)} experts (Wernicke, Broca). Color shows rank percentile (lower = more important) across BLiMP sub-tasks and summary columns. \textbf{[e]} Proportion of dominant BLiMP sub-tasks among the top-2\% units (by each unit’s maximum contribution).}
  \label{fig:fig1}
\end{figure}

\subsection{Phenomenon–Subtype Unit Alignment and p-sweep Robustness}
Having verified subtype separability, we next ask whether subtype-linked units correspond to interpretable linguistic phenomena. We test whether subtype-linked components map to specific linguistic behaviors by aligning their importance with BLiMP phenomenon rankings. For each subtype (Broca, Wernicke), the top 2\% components are selected (dense: neuron groups/feature channels; MoE: experts) and their relative importance across BLiMP subtasks is visualized (Figure~\ref{fig:fig1}). In the dense (non-MoE) baseline, the top $2\%$ subtype-associated components exhibit diffuse importance spread across many BLiMP phenomena, and Broca- and Wernicke-aligned sets yield highly similar profiles (Figure~\ref{fig:fig1}a,b,e). This dispersion suggests that subtype-related signals are encoded in a distributed manner, limiting phenomenon-specific interpretability and making unit-to-function correspondences difficult to localize. In contrast, the MoE model shows a more structured and phenotype-consistent alignment: the top $2\%$ subtype-associated experts form clearer block patterns across BLiMP phenomena (Figure~\ref{fig:fig1}c,d,e), indicating selective sensitivity to particular subsets of linguistic attributes rather than uniform contribution across tasks. This structured selectivity supports the hypothesis that MoE modularity facilitates a more localized mapping from subtype-linked training signals to observable language behaviors, whereas dense models tend to distribute these signals across many units.

Across both architectures, subtype-linked components concentrate their importance in a subset of BLiMP phenomena, most prominently \textit{argument structure}, \textit{ellipsis}, \textit{filler}, \textit{island constraints}, and \textit{$s$-selection}, while showing comparatively weaker associations with \textit{NPI licensing}, \textit{quantifiers}, \textit{binding}, \textit{control/raising}, and aggregate measures (e.g., \textit{semavg}, \textit{synavg}, \textit{synsemavg}). This indicates that the clinically relevant perturbation targets identified by our subtype attribution are not uniformly distributed over linguistic knowledge, but instead disproportionately implicate structure-building and constraint-sensitive phenomena. We also observe subtype differences within this subset: Broca-linked components place relatively greater weight on \textit{argument structure} than Wernicke-linked components, consistent with the view that Broca-like deficits are more tightly tied to production-side structural planning and predicate--argument realization (interpreted here as a model-side correspondence rather than a direct claim about neuroanatomical localization).

\begin{figure}[!t]
  \centering
  \includegraphics[width=\columnwidth]{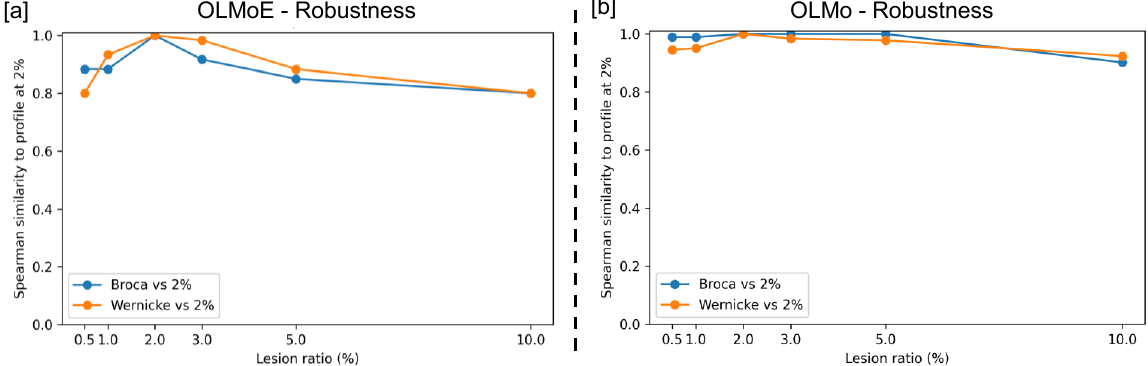}
  \caption{\textbf{Robustness of subtype task profiles around the 2\% lesion choice.} [a] MoE (OLMoE) and [b] dense (OLMo) show Spearman similarity between profiles at lesion ratio $p \in (0.5\%, 1\%, 2\%, 3\%, 5\%, 10\%)$ and the 2\% profile for Broca- and Wernicke-targeted units.}
  \label{fig:fig2}
\end{figure}

Finally, our visualization is robust to the choice of the top-$p$ threshold. Figure~\ref{fig:fig2} shows that phenomenon profiles remain highly stable for $p\in[1\%,5\%]$ (Spearman $\rho>0.85$ relative to the $2\%$ reference) across both architectures and both subtypes. We therefore use $p=2\%$ as a principled compromise that preserves subtype-specific structure while avoiding dilution from lower-signal components. Overall, while both architectures can exhibit phenotype-targeted regressions under a unified component-level analysis, the MoE model concentrates subtype-linked effects into more separable components: perturbing a small subset of experts yields coherent, subtype-consistent shifts that scale with lesion strength. In the dense model, similar regressions likely arise from broader distributed disruption, reflecting weaker modular separability and reduced component-level interpretability.

\subsection{Progressive Lesioning Causes Graded Drops in WAB/AQ}

\begin{figure}[H]
  \centering
  \includegraphics[width=\columnwidth]{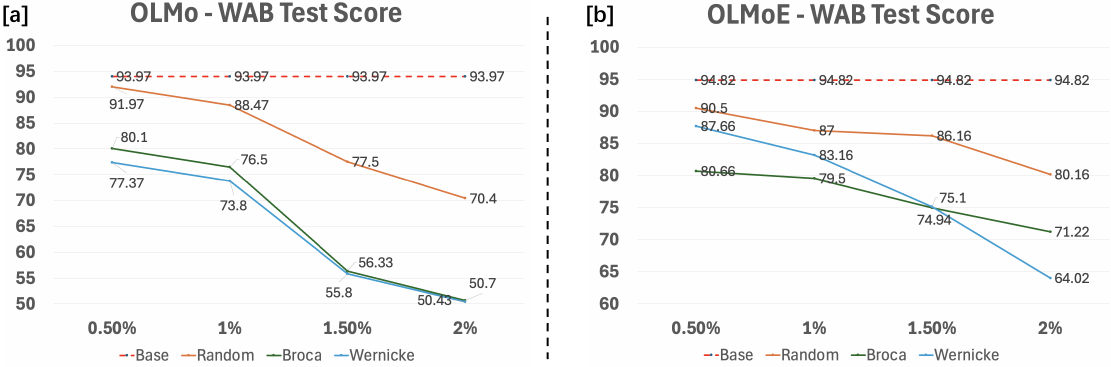}
  \caption{\textbf{WAB Test Score. }Progressive lesioning (Xavier) yields graded clinical degradation on the WAB evaluation (Aphasia Quotient, AQ). [a] Dense OLMo and [b] MoE OLMoE.}
  \label{fig:fig3}
\end{figure}

To test whether phenotype-related unit impairments induce aphasia-like behavior, the most Broca- and Wernicke-associated units are lesioned at increasing ratios ($p\in{0.5,1,1.5,2}\%$) and evaluated with WAB, reporting the Aphasia Quotient (the “WAB test score” in Figure \ref{fig:fig3}). Controls include the intact base model and a random-lesion baseline with the same lesion budget.
In both the non-MoE OLMo model (Figure \ref{fig:fig3}a) and the MoE-based OLMoE model (Figure \ref{fig:fig3}b), the baseline remained stable (93.97–94.82), \textbf{above the WAB-defined threshold score for aphasia (93.8)}. This confirms that the assessment is insensitive to non-lesion-related confounding factors. Conversely, both phenotype-targeted lesions and size-matched random lesions exhibited a clear dose-response relationship: larger perturbations led to a systematic decrease in WAB scores, indicating that our lesion management approach induces graded, clinically interpretable deficits rather than a fragile failure mode.

Crucially, under the same lesion budget, \textbf{phenotype-targeted lesions consistently exhibited greater destructive power than random lesions,} demonstrating that the units selected through phenotype-unit mapping capture functionally and clinically relevant unit components. This effect was most pronounced at high lesion proportions, where random lesions maintained fairly high performance, while phenotype-targeted lesions led to a significant decrease in WAB scores.

Furthermore, we observed a \textbf{separation of model architecture dependencies} between Broca's and Wernicke's functional impairments. In the non-MoE OLMo model (Figure \ref{fig:fig3}a), the dysfunction curves for Broca and Wernicke converge rapidly with lesion ratio p, eventually reaching similar low performance levels. This convergence indicates that in dense Transformer models, the computations supporting these clinical phenotypes are more dispersed and overlapping. This makes it difficult to clearly distinguish subtype-specific dysfunctions under unit ablation. In contrast, the OLMoE model of the MoE architecture (Figure \ref{fig:fig3}b) exhibits more pronounced dysfunction characteristics: as the p-value increases, ablation targeting Wernicke leads to a faster decrease in the WAB score than ablation targeting Broca's (and the gap widens around p = 2\%). This difference aligns with the hypothesis that the \textbf{modular MoE architecture enables more localized and subtype-informative lesion-phenotype mapping, while dense models tend to encode language functions in a more distributed manner}.

\subsection{Qualitative Case Study: Targeted Lesions vs Random and Base}
To complement quantitative results, we compare all the performance of the base model with random lesions and phenotypic targeted lesions (p=2\%) under the same WAB prompt. Table \ref{tab:tb2} shows the results for the WAB prompt (“How are you today?”). The base model produces coherent, socially appropriate responses, and random lesions largely preserve coherence, suggesting that sparse unstructured damage induces only mild noise.

In contrast, phenotype-targeted lesions yield subtype-consistent regressions. Broca-targeted lesions compress production (e.g., OLMoE collapses to minimal affirmations; dense OLMo becomes repetitive), whereas Wernicke-targeted lesions primarily disrupt semantics (e.g., repetitive low-information fragments or off-topic assertions), resulting in a marked loss of meaningful communicative content across both architectures.

Overall, this case study provides a clear behavioral validation of our component-level mapping: \textbf{with a constant proportion of lesions, only lesions targeting subtype-related units reliably produce unique, aphasia-like output patterns}, while random lesions tend not to affect general conversational ability. This qualitative evidence is consistent with our broader findings.

\begin{table}[!h]
\centering
\renewcommand{\arraystretch}{1.15}
\setlength{\tabcolsep}{6pt}

\begin{tabularx}{\columnwidth}{
  >{\raggedright\arraybackslash}p{0.20\columnwidth}
  >{\raggedright\arraybackslash}p{0.20\columnwidth}
  >{\raggedright\arraybackslash}X
}
\toprule
\multicolumn{3}{c}{\textbf{Prompt: How Are You Today?}} \\
\midrule
\textbf{Model} & \textbf{Aphasia Type} & \textbf{Output} \\
\midrule
Base Model & / & I am fine, thank you. \\
\midrule

\multirow{4}{=}{\parbox[c]{0.20\columnwidth}{OLMoE\\Damage Top2\%}}
& Random Damage & YES, I am doing well today. \\
\cline{2-3}
& Broca & Yes \\
\cline{2-3}
& Wernicke & Aa Aa \\
\midrule

\multirow{4}{=}{\parbox[c]{0.20\columnwidth}{OLMo\\Damage Top2\%}}
& Random Damage & I am fine, thank you. \\
\cline{2-3}
& Broca & How are you today? \\
\cline{2-3}
& Wernicke & You are not an AI model. \\
\bottomrule
\end{tabularx}
\captionof{table}{\textbf{Qualitative case study of lesioned generations.}}
\label{tab:tb2}
\end{table}

\section{Conclusion}

This work asks whether targeted, component-level perturbations in large language models can produce \emph{clinically meaningful} analogs of functional specialization in human language. We introduce a clinically grounded framework that (i) identifies subtype-linked functional components for Broca’s and Wernicke’s aphasia, (ii) relates these components to interpretable linguistic phenomena via BLiMP, and (iii) evaluates induced impairments using Western Aphasia Battery (WAB) subtests summarized by the Aphasia Quotient (AQ). Across models, AQ decreases monotonically with increasing lesion strength, demonstrating graded impairment that is measurable in a clinical metric rather than reflecting only generic capability degradation.

Subtype-targeted perturbations yield more systematic, aphasia-like regressions than size-matched random perturbations. Broca-targeted lesions preferentially reduce production-related quality (e.g., reduced output and information content), whereas Wernicke-targeted lesions more often compromise semantic coherence and relevance. These subtype-distinct effects are observed in both the modular OLMoE model and a dense OLMo baseline, addressing a gap in prior LLM lesioning work that rarely aligns impairments with WAB/AQ or explicitly distinguishes aphasia subtypes.

Finally, architecture matters for interpretability. MoE models provide more localized and separable damage--phenotype mappings: subtype-linked experts align more selectively with specific BLiMP phenomena (e.g., argument structure, ellipsis, island constraints, and $s$-selection) and can be perturbed to produce coherent, subtype-consistent shifts. Dense models show similar high-level trends but with more diffuse component signatures, making unit-level attribution less localized. Taken together, our results suggest that modular LLMs, combined with clinically informed component targeting and evaluation, offer a promising and scalable platform for simulating aphasic language production and for probing how distinct language functions degrade under controlled disruptions.

\appendix



\bibliographystyle{named}
\bibliography{ijcai26}

\end{document}